\documentclass[letterpaper]{article} % DO NOT CHANGE THIS
\usepackage{aaai25}  % DO NOT CHANGE THIS
\usepackage{times}  % DO NOT CHANGE THIS
\usepackage{helvet}  % DO NOT CHANGE THIS
\usepackage{courier}  % DO NOT CHANGE THIS
\usepackage[hyphens]{url}  % DO NOT CHANGE THIS
\usepackage{graphicx} % DO NOT CHANGE THIS
\usepackage{natbib}  % DO NOT CHANGE THIS AND DO NOT ADD ANY OPTIONS TO IT
\usepackage{caption} % DO NOT CHANGE THIS AND DO NOT ADD ANY OPTIONS TO IT
\usepackage{algorithm}
\usepackage{algorithmic}
\usepackage{courier}
\usepackage{amsmath} 
\usepackage{amssymb}
\usepackage{subfigure}
\usepackage{multirow}
\usepackage{newfloat}
\usepackage{listings}
\usepackage{color}
\usepackage{threeparttable}
\usepackage{newfloat}
\usepackage{listings}
\DeclareCaptionStyle{ruled}{labelfont=normalfont,labelsep=colon,strut=off} % DO NOT CHANGE THIS
\floatstyle{ruled}
\newfloat{listing}{tb}{lst}{}
\floatname{listing}{Listing}
\title{Towards Accurate Binary Spiking Neural Networks: Learning with Adaptive Gradient Modulation Mechanism}
\author{
    Yu Liang\textsuperscript{\rm 1}, Wenjie Wei\textsuperscript{\rm 1}, Ammar Belatreche\textsuperscript{\rm 2}, \\
    Honglin Cao\textsuperscript{\rm 1}, Zijian Zhou\textsuperscript{\rm 1}, Shuai Wang\textsuperscript{\rm 1}, Malu Zhang\textsuperscript{\rm 1}\thanks{Corresponding author: maluzhang@uestc.edu.cn}, Yang Yang\textsuperscript{\rm 1}
}
\affiliations{
    \textsuperscript{\rm 1}University of Electronic Science and Technology of China\\
    \textsuperscript{\rm 2}Northumbria University

}

\usepackage{bibentry}
\begin{document}

\maketitle

\begin{abstract}
Binary Spiking Neural Networks (BSNNs) inherit the eventdriven paradigm of SNNs, while also adopting the reduced storage burden of binarization techniques. These distinct advantages grant BSNNs lightweight and energy-efficient characteristics, rendering them ideal for deployment on resource-constrained edge devices. However, due to the binary synaptic weights and non-differentiable spike function, effectively training BSNNs remains an open question. In this paper, we conduct an in-depth analysis of the challenge for BSNN learning, namely the frequent weight sign flipping problem. To mitigate this issue, we propose an Adaptive Gradient Modulation Mechanism (AGMM), which is designed to reduce the frequency of weight sign flipping by adaptively adjusting the gradients during the learning process. The proposed AGMM can enable BSNNs to achieve faster convergence speed and higher accuracy, effectively narrowing the gap between BSNNs and their full-precision equivalents. We validate AGMM on both static and neuromorphic datasets, and results indicate that it achieves state-of-the-art results among BSNNs. This work substantially reduces storage demands and enhances SNNs' inherent energy efficiency, making them highly feasible for resource-constrained environments.
\end{abstract}

\section{Introduction}

Deep Neural Networks (DNNs) have achieved significant breakthroughs in various fields, such as computer vision~\cite{he2016identity,liu2024sora}, natural language processing~\cite{vaswani2017attention,achiam2023gpt}, and speech processing~\cite{wu2018spiking,pan2021multi}.
However, these successes rely on complex model architectures and large model parameters, making their deployment on resource-limited devices highly challenging~\cite{qin2024accurate}.
In contrast, brain-inspired spiking neural networks (SNNs) emulate the spike generation mechanism of biological neurons, utilizing sparse binary spikes as the basic unit for information transmission~\cite{ghosh2009spiking}.
Due to their sparse and spike-driven computational paradigm, SNNs have been regarded as the energy-efficient alternative to traditional DNNs~\cite{davies2018loihi}.
Nonetheless, despite their energy efficiency, SNNs often struggle to achieve satisfactory performance when dealing with complex tasks \cite{guo2023direct}.

To ensure the competitive performance of SNNs on complex tasks, the SNN community is committed to developing effective training strategies and large-scale architectures~\cite{wu2021progressive,yao2024spike,guo2022real}, often neglecting the inherent energy efficiency trait of SNNs. 
This negligence increases the difficulty of deploying SNNs on resource-constrained edge computing environments~\cite{liu2023low,guo2022recdis}.
Therefore, a deeper understanding of the energy efficiency advantage of SNNs is crucial to enable their widespread deployment, especially in resource-limited scenarios \cite{wei2024event}. 
In recent years, some studies have conducted preliminary exploration on compressing SNNs, including techniques such as pruning~\cite{li2024towards,shi2023towards}, neural architecture search~\cite{liu2024lite,yan2024efficient}, and quantization~\cite{hu2024bitsnns}.

Among these compression techniques, quantization is a promising solution, which reduces computational and storage demands by representing weights with low bit width~\cite{gong2014compressing,li2019additive}.
The most extreme method within this domain is binarization, which compresses networks by reducing the parameter bit width to a mere 1-bit~\cite{rastegari2016xnor,qin2022bibert,gholami2022survey}.
Incorporating this binarization into SNNs allows computationally intensive convolutional operations to be replaced with efficient bitwise operations. This transformation significantly reduces both computation and storage requirements, making networks more hardware-friendly and energy-efficient.
However, despite the significant advantages in binarizing SNNs, these approaches struggle to scale to complex tasks when compared to full-precision SNNs (FP-SNNs)~\cite{yinmint}.
There is no extensive research on the underlying reason behind this critical challenge.

In this paper, we reveal that due to the two possible values of their weights (-1 and +1), binary SNNs (BSNNs) face a more severe challenge of weight sign flipping during the learning process.
This issue causes the performance of BSNNs to lag behind that of FP-SNNs equivalents, limiting their application in complex scenarios.
To address this issue, we propose an adaptive gradient modulation mechanism (AGMM) to adaptively regulate the gradient magnitude during the learning process.
By integrating AGMM, BSNNs can mitigate the frequent weight sign flipping problem, leading to faster convergence and higher accuracy.
% \textcolor{blue}{rewrite this paragraph} 
We summarize the main contributions of this paper as follows: 
\begin{itemize}
\item We provide a comprehensive analysis of the difficulty in BSNNs learning, i.e., the frequent weight sign flipping problem. Furthermore, we reveal that the frequency of flipping is proportional to the mean and variance of gradients via rigorous reasoning.
\item We propose an Adaptive Gradient Modulation Mechanism (AGMM), aiming to adjust the gradient magnitude in the learning process to solve the frequent weight sign flipping problem. This mechanism effectively narrows the gap between BSNNs and their FP-SNN equivalents.
\item Extensive experiments conducted on both static and neuromorphic datasets demonstrate that our method achieves state-of-the-art performance when compared with other existing BSNNs approaches.
Moreover, we thoroughly validate the effectiveness and efficiency of the AGMM.
\end{itemize}

\section{Related Work}
% SNN BNN BSNN
\subsection{Spiking Neural Network}

Brain-inspired SNNs have emerged as a new computing paradigm \cite{zhang2019mpd,tang2024neuromorphic}. Unlike traditional DNNs, SNNs perform computations using binary spikes over time, making them especially suitable for energy-efficient computing applications \cite{wang2024global,wei2023temporal}.
Given the non-differentiability of the spike generation function, the SNN research community has devoted considerable efforts to developing effective training algorithms. These approaches can be broadly classified into two categories: conversion-based methods and direct learning algorithms.
Conversion methods may not fully exploit the temporal processing capabilities of SNNs and often require multiple steps for accurate inference \cite{roy2019scaling,wang2020deep}. In contrast, direct learning methods can achieve high accuracy with low-latency inference \cite{wu2018spatio,wu2019direct,wang2024ternary}. 
Considering activation values in SNNs are binary, further quantizing the weights to binary would allow computation-intensive convolution operations to be executed using simple bitwise operations. This would further enhance the inherent efficiency advantages of SNNs.

\subsection{Binary Neural Network}
% BNN 权重是二值，xxx
As an extreme form of network quantization, binary neural networks (BNNs) constrain both weights and activations to -1 and +1, substantially reducing memory storage and computational overhead. However, BNNs experience substantial performance degradation compared to their full-precision counterparts. Subsequent research has focused on narrowing this performance gap between BNNs and full-precision neural networks. Several approaches have been proposed to address this issue.
XNOR-Net~\cite{rastegari2016xnor} employs a deterministic binarization scheme and minimizes quantization error by incorporating scaling factors in each layer.
Bi-Real~\cite{liu2018bi} explores piecewise polynomial functions and network architectures for quantization. Previous work has also focused on constraining the distribution of weights or activations in BNNs~\cite{xu2021recu}. 
Building on these advancements in the binary domain, binarization methods in SNNs remain a promising area for exploration.

\subsection{Binarization techniques in SNNs}

Several researchers have combined SNNs with binarization techniques.
For example,~\cite{qiao2021direct} employ the surrogate gradient (SG) method to train BSNNs directly. 
\cite{jang2021bisnn} propose a novel Bayesian-based algorithm for BSNNs, which shows significant advantages in accuracy and calibration compared to SG methods. 
\cite{kheradpisheh2022bs4nn} integrate binarization into temporal-coded SNNs, where each neuron emits at most one spike, offering substantial energy benefits. 
Recently, \cite{wei2024q} take inspiration from information theory and propose a weight-spike dual regulation (WS-DR) method to enhance the information capacity of BSNNs, achieving significant performance improvements. 
However, current BSNNs continue to encounter challenges when scaling to complex datasets such as ImageNet.
Therefore, it is essential to investigate the challenges inherent in the BSNN training process and to bridge the performance gap between BSNNs and FP-SNNs.

\section{Preliminary}
\paragraph{Leaky Integrate-and-Fire model.} 
Various spiking neuron models have been proposed~\cite{Hodgkin_Huxley,zhang2021rectified}. In this paper, we select the Leaky Integrate-and-Fire (LIF) model due to its computational efficiency. Typically, the Euler method is employed  to discretize the dynamic equations of LIF models~\cite{wu2018spatio}, resulting in the following iterative form: 
\begin{align}
\label{u}
U^{n}[t]=\tau U^{n}[t-1]+ X^{n}[t],
\end{align}
where \(\tau\) is the constant leaky factor, \(U^{n}[t]\) represents the membrane potential of neurons in layer \(n\) at time \(t\), and \(X^n[t]\in \mathbb{R}^{(C, H, W)}\) is their input current. \(X^n[t]\) integrates inputs from all presynaptic neurons, described as:
\begin{align}
\label{current}
X^{n}[t] = \mathcal{BN}\left( W^n S^{n-1}[t]\right),
\end{align}
where $\mathcal{BN}(\cdot)$ denotes the batch normalization operation, $W^n$ is the weights of the $n$-th convolutional layer, and $S^{n-1}[t]$ is the output spikes of neurons in layer $n-1$ at time $t$.
When the membrane potential $U^{n}[t]$ reaches the firing threshold $V_{th}$, the neuron will generate a spike, defined as:
\begin{align}
S^{n}[t]
=\left\{\begin{matrix}
\;1, &\! \text{if}~~{U}^{n}[t] \geq V_{th},\\ 
\;0, &\! \text{otherwise}.\\
\end{matrix}\right.
\label{spikefunc}
\end{align}
After spike emission, the hard reset function is invoked, mathematically defined as:
\begin{align}
U^{n}[t]=U^{n}[t]\cdot \left(1-S^{n}[t]\right).
\end{align}
Clearly, $U^n[t]$ is reset to zero upon spike emission while
remaining unchanged if no spike occurs.

\subsubsection{Weight binarization}
To achieve significant model compression, BSNNs adopt the sign function to convert $W^n$ in Eq.~\ref{current} into binary representations.
Typically, the scaling factor $\gamma$ is introduced to match the distribution of float-point weights, thus minimizing the quantization error.
Therefore, the binarization on $\omega\in W^n$ can be formulated as:
\begin{align}
    {\omega_b}=\gamma \cdot \mathrm{sign}({\omega})=\left\{\begin{matrix}
    -\gamma, & \text{if}~~\omega<0,\\
    +\gamma, & \text{otherwise,}\end{matrix}\right.
    \label{sign}
\end{align}
where $\omega_b$ is a binary weight (\(\omega_b\in W^n_b\)), and $\gamma$ is computed as the average of the absolute value of weights in each channel~\cite{rastegari2016xnor}.
To solve the non-differentiability of $\operatorname{sign}(\cdot)$, the straight-through estimator (STE) is employed to approximate the gradient of binary weights~\cite{bengio2013estimating}, i.e., $\frac{\partial\operatorname{sign(\cdot)}}{\partial W^{n}}=1_{|W^{n}|\leq1}$.

\subsubsection{Learning algorithm}
In order to train SNNs successfully, we use the spatial-temporal backpropagation (STBP) algorithm~\cite{wu2018spatio}, where the gradient of the loss function $\mathcal{L}$ to weight $W_b^n$ can be formulated as:
\begin{equation}
\begin{aligned}
    \frac{\partial \mathcal{L}}{\partial W_b^{n}} &= \sum_t \left( \frac{\partial \mathcal{L}}{\partial S^{n}[t]} \frac{\partial S^{n}[t]}{\partial U^{n}[t]} + \right. \\
    &\quad \left. \frac{\partial \mathcal{L}}{\partial U^{n}[t+1]} \frac{\partial U^{n}[t+1]}{\partial U^{n}[t]} \right) \frac{\partial U^{n}[t]}{\partial W_b^n},
\end{aligned}
\end{equation}

where $\frac{\partial S^{n}[t]}{\partial U^{n}[t]}$ is the derivative of the non-differentiable spike step function. Therefore, the surrogate gradient is employed to replace this gradient. In this paper, we use the triangular-shaped function, which is mathematically defined as:
\begin{align}
    \frac{\partial S^{n}[t]}{\partial U^{n}[t]}=\max\left(0,\beta-\left|U^{n}[t]-V_{th}\right|\right),
\end{align}
where $\beta$ is a hyperparameter that controls the gradient range.
\begin{figure}[tbp]
    \centering
    \subfigure[]{\includegraphics[width=0.49\linewidth]{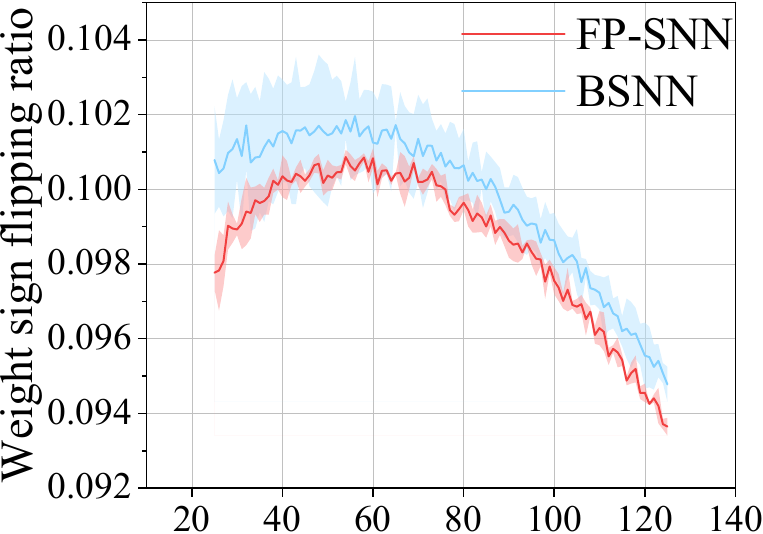}\label{fig:fp_base_flip}}
    \subfigure[]{\includegraphics[width=0.50\linewidth]{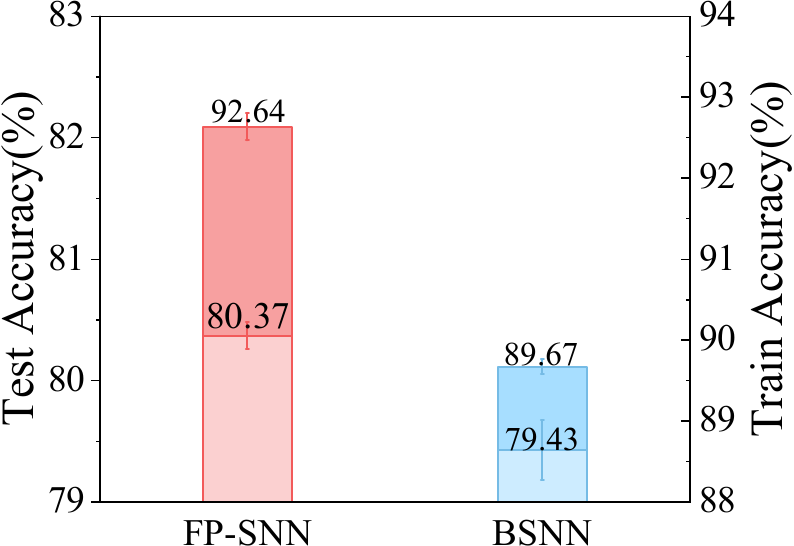}\label{fig:fp_base_acc}}
    \caption{Weight sign flipping ratio and performance comparison between FP-SNN and vanilla BSNN. The darker color represents the training accuracy, while the lighter color indicates the test accuracy.}
    \label{fig:fP_base}
\end{figure}
\begin{figure}[tbp]
    \centering
    \subfigure[FP-SNN]{\includegraphics[width=0.49\linewidth]{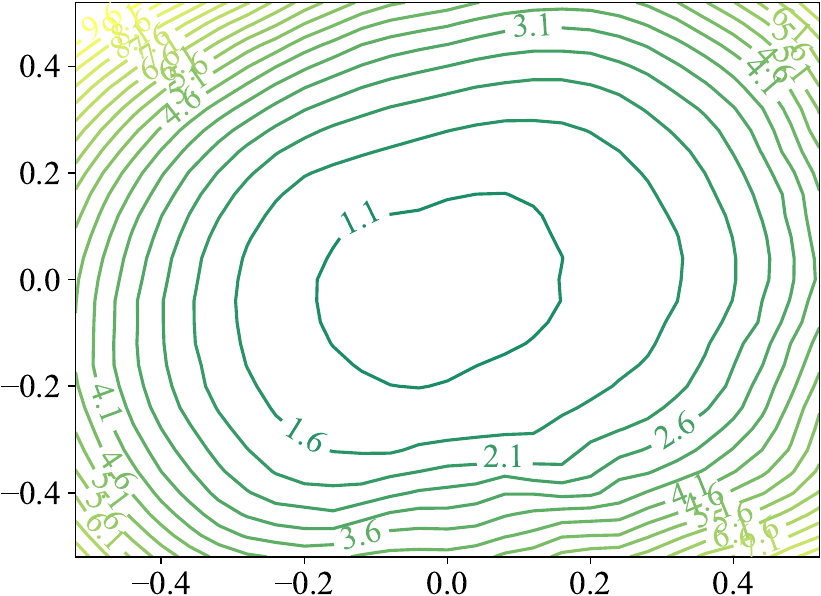}}
    \subfigure[Vanilla BSNN]{\includegraphics[width=0.49\linewidth]{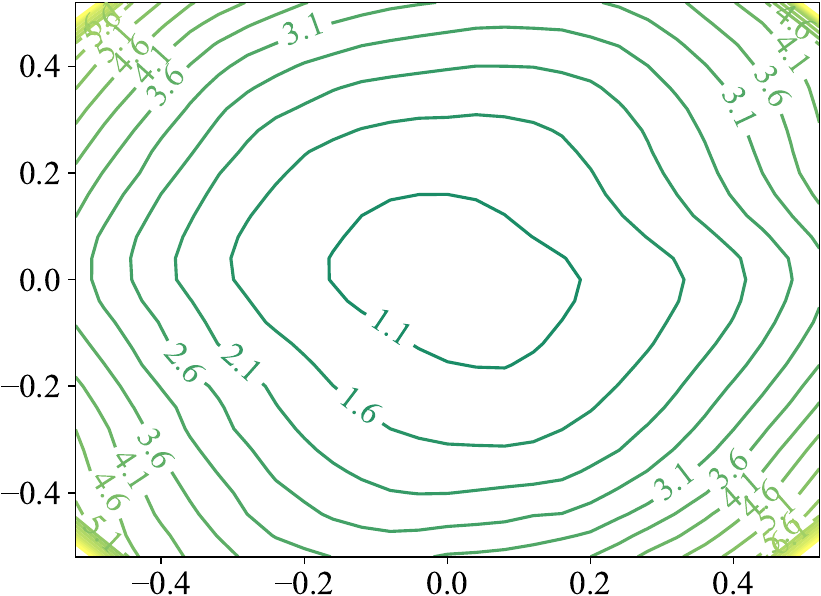}}
    \caption{The loss landscape of FP-SNN and vanilla BSNN. }
    \label{fig:fp_loss}
\end{figure}

\begin{figure*}
    \centering
    \includegraphics[width=0.9\linewidth]{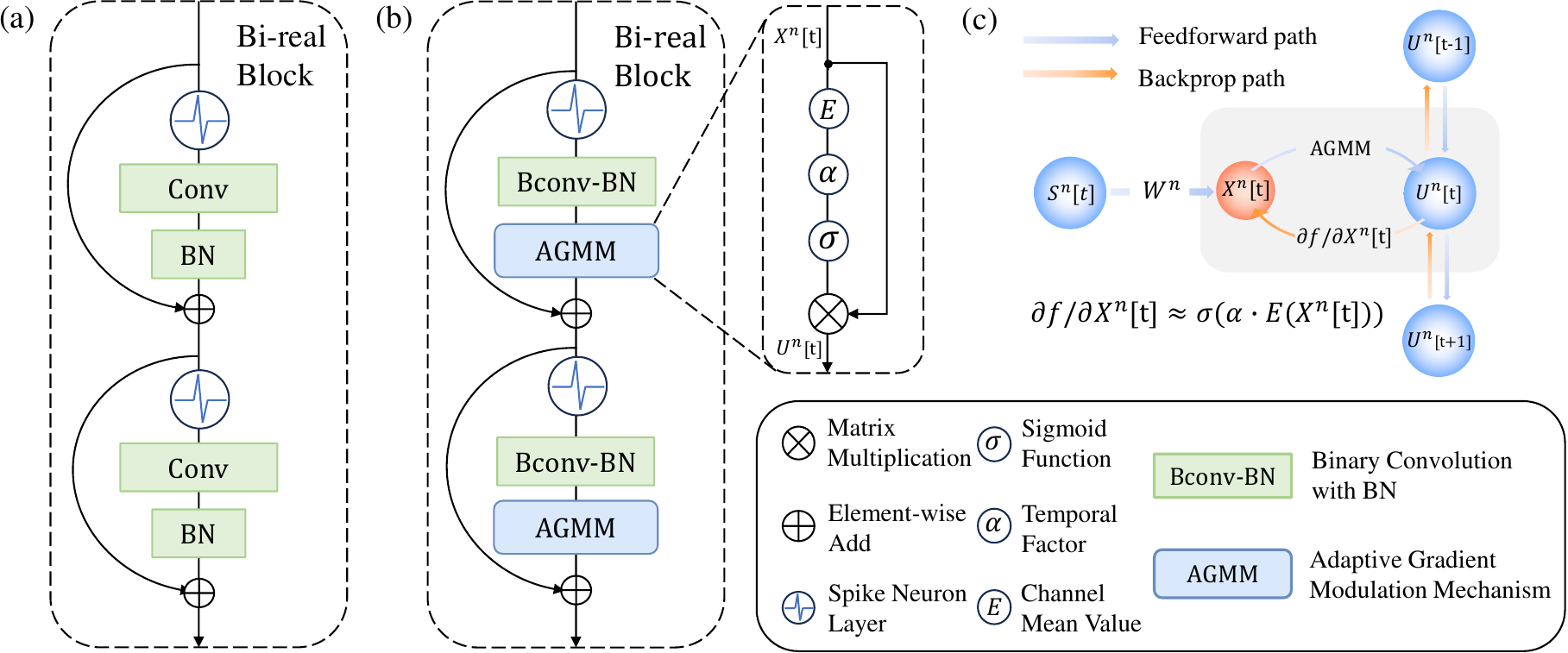}
    \caption{(a) The network architecture of FP-SNNs, which is inherited from the work of Bi-Real Net~\cite{liu2018bi}. (b) The proposed AGMM module in BSNNs, where each Bconv-BN module is followed by an AGMM module. (c) The backpropagation computation graph of the proposed AGMM.}
    \label{fig:backbone}
\end{figure*}
\section{Method}
In this section, we first conduct an in-depth analysis of the frequent weight sign flipping problem in BSNNs. Subsequently, we propose an AGMM method to adaptively adjust the gradient magnitude in the BSNN learning process, thereby addressing the aforementioned issue.
\subsection{Problem analysis}

Compared to FP-SNNs, BSNNs offer substantial advantages in energy efficiency, making them highly suitable for deployment on edge devices. However, scaling BSNNs to more complex tasks or datasets is still a major challenge.
This is primarily attributed to the frequent weight sign flipping problem in the learning process of BSNNs.

Specifically, FP-SNNs update the weight $\omega$ in a continuous space, allowing fine-grained adjustments. In contrast, BSNNs update $\omega$ in a discrete space, namely -1 and 1, essentially flipping the sign. While an appropriate flipping frequency can help achieve stable convergence and high-performance networks, BSNNs suffer from more frequent weight sign flipping compared to FP-SNNs. To measure this difference, we adopt the metric proposed by \cite{helwegen2019latent} to calculate the weight sign flipping ratio during the learning process. This metric is defined as:
\begin{align}
\label{flip-flop}
\pi_e = \frac{\sum_{n}\operatorname{sum}(\text{sign}(W_{e}^{n}) \oplus \text{sign}(W_{e+1}^{n}))}{\sum_{n} \left | W_{e}^{n} \right |} ,
\end{align}
where $W_{e}^{n}$ is the weight in layer $n$ at epoch $e$, $\operatorname{sum}(\cdot)$ is the sum function, and $\left | W_{e}^{n} \right |$ is  the number of entries in $ W_{e}^{n}$.

We record the flipping ratio and performance of FP-SNN and BSNN throughout the overall training process on CIFAR-100 with the ResNet-19 and plot them in Fig.~\ref{fig:fP_base}.
From Fig.~\ref{fig:fp_base_flip}, where the x-axis denotes the number of epochs, we observe that BSNN consistently exhibits a higher flipping ratio compared to FP-SNN throughout the training process. Besides, we can also find that the flipping ratio is gradually decreased as the BSNN converges.
The performance results for FP-SNN and BSNN are shown in Fig.~\ref{fig:fp_base_acc}, revealing a significant accuracy gap between FP-SNN and BSNN. 
To demonstrate the training difference between FP-SNN and BSNN, we also analyze their loss lanscapes\cite{li2018visualizing}. According to Fig.~\ref{fig:fp_loss}, in contrast to FP-SNN, the loss for BSNN increases sharply around the borders of the loss landscape. This sharpness of the network's loss landscape correlates well with its generalization error. This result shows that FP-SNN outperforms BSNN in terms of generalization.

Given the frequent weight sign flipping issue occurring during BSNNs training, it is obvious that balancing the frequency of weight flipping is crucial for better convergence and performance. Fortunately, the frequency of weight sign flipping is positively correlated with the mean and variance of gradients. 
Next, we conduct a rigorous analysis of this.
\paragraph{Proposition 1.} \textit{The frequency of weight sign flipping is positively correlated with the mean and variance of gradients. Specifically, the larger the mean ($\mu$) or variance ($\sigma^2$) of gradients, the higher the frequency of weight sign flipping.} 
\paragraph{Proof 1.}\label{proof1} Consider a weight $\omega$ for analysis. According to gradient descent optimization, its update can be described as:
\begin{align}
\omega_{e+1}=\omega_e-\eta \nabla \mathcal{L}(\omega_e),
\end{align}
where $e$ is the epoch, $\eta$ is the learning rate, and $\nabla  \mathcal{L}(\omega_e)$ is the gradient of the loss function to the weight $\omega$.
Based on Eq.~\ref{flip-flop}, we define the weight sign flipping event $A_e$ as $\text{sign}(\omega_e)\neq \text{sign}(\omega_{e+1})$.

Assume $\omega_e > 0$ (the case for $\omega_e < 0$ is similar), then the condition for $A_e$ to occur is $\omega_e - \eta \nabla \mathcal{L}(\omega_e) < 0$.
Generally, gradients in neural networks follow a normal distribution \cite{glorot2010understanding,he2015delving,zhang2018three}, i.e., $\nabla \mathcal{L}(\omega_e)\sim N(\mu, \sigma^2)$. Under this distribution, the probability of event $A_e$ can be calculated as:
\begin{align}
\begin{split}
\label{eq:p_ae}
P(A_e) &= P(\nabla \mathcal{L}(\omega_e) > \frac{\omega_e}{\eta}) \\
% 1 - \Phi\left(\frac{\frac{w_k}{\eta} - \mu}{\sigma}\right)
&=1-F(\frac{\omega_e}{\eta};\mu,\sigma) \\
&=1-\frac1{\sigma\sqrt{2\pi}}\int_{-\infty}^{\frac{\omega_e}{\eta}}\exp\left(-\frac{(t-\mu)^2}{2\sigma^2}\right)dt,
\end{split}
\end{align}
where $F(\cdot)$ is the cumulative distribution function of $\nabla \mathcal{L}(\omega_e)$.
From Eq.~\ref{eq:p_ae}, it can be observed that $P(A_e)$ is positively correlated with $\mu$ and $\sigma^2$. 
Therefore, in BSNNs, the frequent weight sign flipping problem can be alleviated by reducing $\mu$ and $\sigma^2$ of the gradients in the backward process.

\subsection{Adaptive Gradient Modulation Mechanism}
Based on the above analysis, we propose an Adaptive Gradient Modulation Mechanism (AGMM), which adaptively reduces gradient magnitudes during the learning process, as shown in Fig.~\ref{fig:backbone}. In the following, we analyze how AGMM works from its forward and backward propagation.
\subsubsection{Forward propagation}
\begin{figure}
    \centering
    \includegraphics[width=0.9\linewidth]{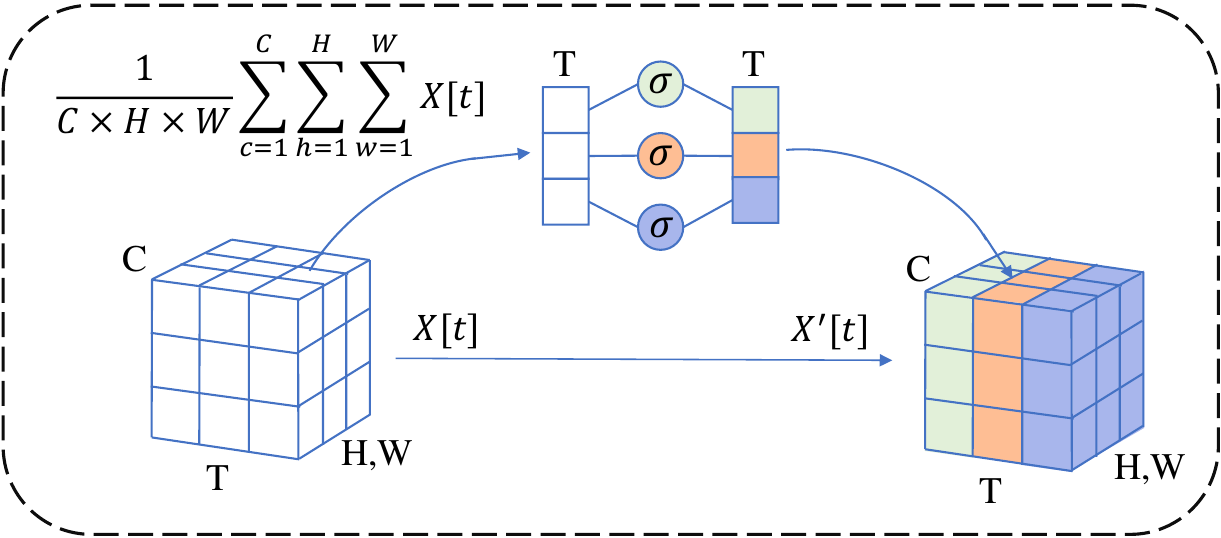}
    \caption{Diagram of AGMM. $X[t]$ is computed by Bconv-BN module, where T denotes the time step, C denotes the channel, and H, W represents the spatial resolution.}
    \label{fig:agmm}
\end{figure}
In the forward propagation, the proposed AGMM is performed on the spatial feature map $X^n[t]$, which is calculated by the Bconv-BN module, as shown in Fig.~\ref{fig:agmm}. For simplicity, we ignore the layer index $n$ in the following analysis. The AGMM is mathematically defined as follows:
\begin{equation}
\begin{aligned}
\hspace{1.3cm} X'[t]&= \sigma (E(X[t]))\cdot X[t],  &\\
E(X[t])&=\frac{\alpha[t]}{C\times H \times W}\sum_{c=1}^{C}\sum_{h=1}^{H}\sum_{w=1}^{W}X[t],
\end{aligned}
\end{equation}
where $X'[t]$ denotes the output feature map calculated by AGMM, $\sigma(\cdot)$ is the sigmoid function, and $\alpha[t]$ is the trainable temporal-wise factor. 
During the forward pass, AGMM computes the time-scaled average $E(X[t])$ with the temporal factor $\alpha[t]$, and modulates the feature map via a sigmoid function to generate the enhanced output $X'[t]$.

AGMM preserves the network's spike-driven characteristics despite introducing negligible parameters. Furthermore, the sigmoid function of AGMM lowers the firing rate of the network, which in turn lowers synaptic operations (SOPs). The Experiments section offers comprehensive evidence supporting these claims.
\subsubsection{Backward propagation}

To better understand the effectiveness of AGMM in training, we analyze the gradient magnitude throughout the network and provide a thorough explanation of AGMM's backpropagation mechanism. 
In AGMM-BSNN, the derivative of the loss function with respect to the binary weight $\omega_b$ is described as:
\begin{align}
\label{weight backward}
\begin{aligned}
\frac{\partial\mathcal{L_{\text{AGMM}}}}{\partial \omega_b} 
&=\sum_{t=1}^{T}\frac{\partial\mathcal{L_\text{AGMM}}}{\partial X'[t]}\frac{\partial X'[t]}{\partial X[t]}\frac{\partial X[t]}{\partial \omega_b}.
\end{aligned}
\end{align}
Compared to that of vanilla BSNN, this derivative includes an additional term $\partial X'[t]/ \partial X[t]$. In the following analysis, we primarily focus on the additional term. 
By applying the chain rule, this additional term can be expressed as follows:
\begin{equation}
\label{atmm backward}
\begin{split}
\quad\frac{\partial X'[t]}{\partial X[t]} 
&= \frac{\partial {( \sigma (E(X[t])\cdot X[t]))}}{\partial X[t]} \\
= \sigma &(E(X[t]))\frac{\partial X[t]}{\partial X[t]}+X[t]\frac{\partial \sigma( E(X[t]))}{\partial X[t]}.
\end{split}
\end{equation}
The first term in Eq.~\ref{atmm backward} is simplified to $\sigma (E(X[t]))$, while the second term can be expanded using the derivative of the sigmoid function, i.e., $\sigma'(x) = \sigma(x)(1-\sigma(x))$. Therefore, the derivative in the second term of Eq.~\ref{atmm backward} can be rewritten as: 
\begin{equation}
\begin{split}
\frac{\partial \sigma( E(X[t]))}{\partial X[t]}=\frac{\partial E(X[t])}{\partial X[t]}\sigma (E(X[t]))(1-\sigma (E(X[t]))),
\end{split}
\label{14}
\end{equation}
where $\partial E(X[t])/\partial X[t]\!=\!\frac{\alpha[t]}{C\times H\times W}$ and $\sigma ( E(X[t]))(1-\sigma (E(X[t])) \in [0,0.25]$. Therefore, the numerator of the term in Eq. \ref{14} is substantially smaller than the denominator. Building upon this, the second term of Eq.~\ref{atmm backward} can be neglected. Consequently, it can be rewritten as a simplified form:
\begin{equation}
\begin{aligned}
\label{eq:approx}
\frac{\partial X'[t]}{\partial X[t]} \approx \sigma (E(X[t])).
\end{aligned}
\end{equation}

\begin{table*}[t]
\centering
\def\arraystretch{1.2}

\begin{tabular}{llllll} 
\hline
Dataset & Method & Architecture & Learning  & Timestep & Accuracy \\ \hline
\multirow{7}{*}{CIFAR-10} 
&~\cite{wang2020deep}  & 6Conv3FC & ANN2SNN  & 100  & 90.19\%  \\
&~\cite{yoo2023cbp} & VGG16   & ANN2SNN  & 32    & 91.51\%  \\&~\cite{deng2021comprehensive} & 7Conv3FC   & Direct train & 8  & 89.01\%  \\& ~\cite{pei2023albsnn}   & 5Conv1FC  & Direct train & 1   & 92.12\%  \\ & ~\cite{wei2024q} & ResNet-19  & Direct train & 2   & 95.54\%  \\  \cline{2-6} 
 & \multirow{2}{*}{\textbf{AGMM}} & \multirow{2}{*}{ResNet-19} & \multirow{2}{*}{Direct train} & 2 & \textbf{96.13\%} \\
 & & & & 4 &\textbf{96.33\%}\\
\hline
\multirow{6}{*}{CIFAR-100}
% &~\cite{roy2019scaling}& VGG16 & ANN2SNN & - & 54.44\% \\
 &~\cite{lu2020exploring} & VGG15 & ANN2SNN & 400 & 62.07\% \\
 &~\cite{wang2020deep}& 6Conv2FC & ANN2SNN & 300 & 62.02\% \\
 &~\cite{yoo2023cbp}& VGG16 & ANN2SNN & 32 & 66.53\% \\
 &~\cite{deng2021comprehensive}& 7Conv3FC & Direct train & 8 & 55.95\% \\
 &~\cite{pei2023albsnn}& 6Conv1FC & Direct train & 1 & 69.55\% \\ 
  &~\cite{wei2024q}& ResNet-19 & Direct train & 2 & 78.82\% \\ 
\cline{2-6}
  & \multirow{2}{*}{\textbf{AGMM}} & \multirow{2}{*}{ResNet-19} & \multirow{2}{*}{Direct train} & 2 & \textbf{80.25\%} \\
 & & & & 4 &\textbf{80.70\%}\\
\hline
\multirow{4}{*}{ImageNet} & \multirow{2}{*}{~\cite{yoo2023cbp}} & ResNet-18 & \multirow{2}{*}{Direct train} & \multirow{2}{*}{4} & 54.34\% \\
 &               & ResNet-34  &              &   & 60.10\%          \\
 &~\cite{hu2024bitsnns}  & ResNet-18 & Direct train & 1 & 62.06\%          \\ \cline{2-6} 
 & \textbf{AGMM} & ResNet-18 & Direct train & 4 & \textbf{64.67\%} \\ \hline
\multirow{5}{*}{DVS-Gesture} 
 &~\cite{pei2023albsnn}& 5Conv1FC & Direct train & 20 & 94.63\% \\
 &~\cite{qiao2021direct}& 2Conv2FC & Direct train & 150 & 97.57\% \\
 &~\cite{yoo2023cbp}& 15Conv1FC & Direct train & 16 & 97.57\% \\ 
  &~\cite{wei2024q}& VGGSNN & Direct train & 16 & 97.92\% \\ 
\cline{2-6}
 & \textbf{\textbf{AGMM}} & VGGSNN & Direct train & 16 & \textbf{97.92\%} \\ 
\hline
\multirow{5}{*}{DVS-CIFAR10} 
 &~\cite{qiao2021direct}& 2Conv2FC & Direct train & 25 & 62.10\% \\
 &~\cite{pei2023albsnn}& 5Conv1FC & Direct train & 10 & 68.98\% \\
 &~\cite{yoo2023cbp}& 16Conv1FC & Direct train & 16 & 74.70\% \\ 
  &~\cite{wei2024q}& VGGSNN & Direct train & 10 & 81.60\% \\ 
\cline{2-6}
 & \textbf{\textbf{AGMM}} & VGGSNN & Direct train & 10 & \textbf{82.40\%} \\
\hline
\end{tabular}
% \begin{tablenotes} 
% \end{tablenotes} 
\caption{Performance comparison on both static and neuromorphic datasets.}
\label{tab:acc}
\end{table*}

Building on the above analysis, the gradients of the binary weight $\omega_b$ in a BSNN with AGMM can be expressed as follows:
\begin{align}
\nabla \mathcal{L}_{\text{AGMM}}(\omega_b) \approx \sum_{t=1}^{T} \nabla \mathcal{L}_{t}(\omega_b) \cdot \sigma( E(X[t])),
\end{align}
where the gradient distribution \( \nabla \mathcal{L}_t(\omega_b) \) at time \( t \) follows a normal distribution \( N(\mu_t, \sigma_t^2) \), and \( \nabla \mathcal{L_\text{AGMM}}(\omega_b) \) follows a normal distribution \( N(\mu, \sigma^2) \). As a result, both the mean and variance of the gradients at time \( t \) are scaled by \( \sigma(E(X[t])) \).
Given that $0<\sigma(\cdot)<1$, this scaling reduces both the mean and variance of the gradient.
By combining with Proof 1, we can derive:
\begin{align}
P_{\text{AGMM}}(A_e) < P(A_e).
\end{align}
Therefore, using the AGMM can effectively reduce the probability of frequent weight sign flipping. The sigmoid function $\sigma(\cdot)$ reduces the original gradient values while the term $E(X[t])$ determines the extent of gradient reduction based on the time step. 
By combining these two characteristics in backpropagation, AGMM can adaptively reduce the gradient magnitude to mitigate the problem of frequent weight sign flipping in BSNNs.

\section{Experiments}

\subsection{Implementation details}
The experimental evaluation focuses on image classification tasks across two categories of datasets: static and neuromorphic. The static datasets include CIFAR-10, CIFAR-100 and ImageNet while the neuromorphic datasets comprise DVS-Gesture and DVS-CIFAR10. These datasets are highly significant in the fields of machine learning and neuromorphic computing, serving as standard benchmarks for evaluating a variety of methods.
Following previous studies~\cite{rastegari2016xnor}, we maintain full-precision weights in the final fully-connected layer and the first convolutional layer. We evaluate our AGMM using ResNet-19 on CIFAR-10 and CIFAR-100 with 400 epochs, and ResNet-18 on ImageNet with 300 epochs. For ResNet architecture, we use the double skip connections~\cite{liu2018bi}. For neuromorphic datasets, we implement the structure of VGGSNN commonly used in SNNs~\cite{deng2022temporal}.
We employ SGD as the optimizer, coupled with a cosine annealing schedule for learning rate adjustment. 

\subsection{Comparison with related work}
In this section, we evaluate the efficacy and efficiency of the proposed AGMM by comparing its performance and model size with existing quantized SNN approaches. 

We demonstrate the efficacy of AGMM-BSNN by comparing it with other quantized methods.
As shown in Tab.~\ref{tab:acc}, AGMM-BSNN consistently outperforms all quantized methods across various datasets.
On the CIFAR-10 dataset, AGMM-BSNN achieves 96.13\% and 96.33\% accuracy with 2 and 4 timesteps respectively, outperforming other existing approaches. 
This performance exceeds the previous best result, Q-SNN \cite{wei2024q}, by 0.59\% under the same experimental settings.
On CIFAR-100, AGMM-BSNN also demonstrates superior performance, achieving the accuracy of 80.25\% and 80.70\% with 2 and 4 timesteps. This surpasses the state-of-the-art (SOTA) performance Q-SNN by 1.43\% with the same timestep and structure.
For ImageNet, AGMM-BSNN outperforms the reported best result Bit-SNN\cite{hu2024bitsnns}, with an accuracy of 64.67\%. 
On neuromorphic datasets, AGMM-BSNN matches the SOTA performance of Q-SNN on DVS-Gesture and exceeds it on DVS-CIFAR10, obtaining 82.40\% accuracy on DVS-CIFAR10.

\begin{figure}
    \centering    
    \includegraphics[width=0.90\linewidth]{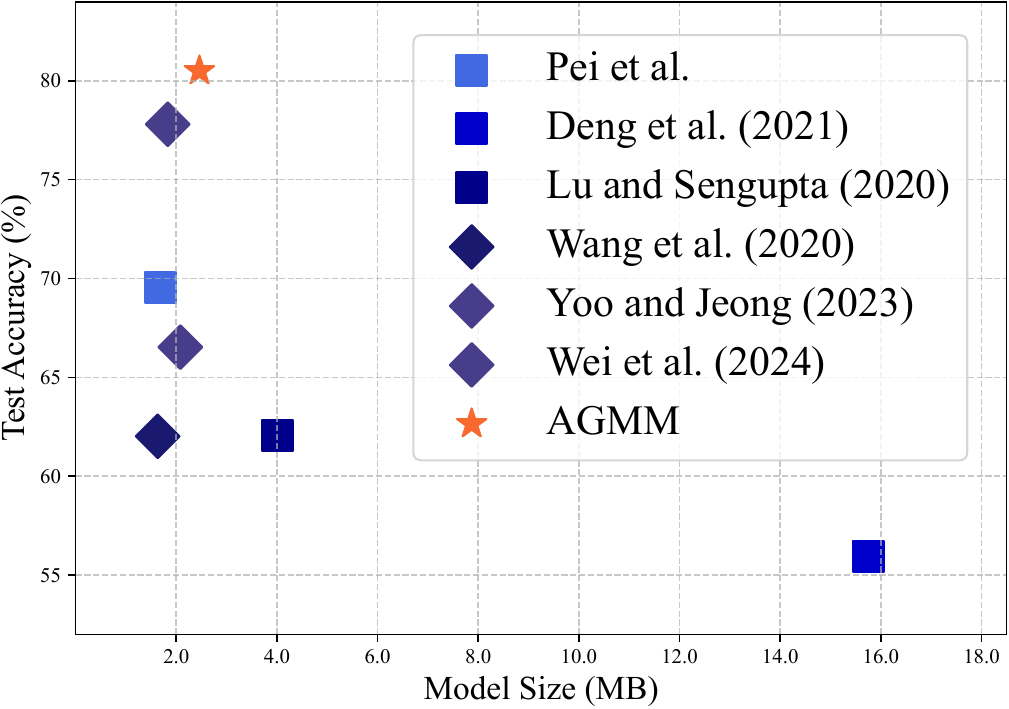}
    \caption{Comparison of test accuracy and model size  of AGMM and other methods on CIFAR-100.}
    \label{fig:modelSize}
\end{figure}

Our model achieves superior performance while maintaining computational efficiency comparable to existing approaches. We analyze the AGMM-BSNN model size in comparison to existing quantized SNN methods on CIFAR-100, with the results illustrated in Fig.~\ref{fig:modelSize}. AGMM-BSNN achieves 80.25\% accuracy with only 2.46MB, outperforming Q-SNN~\cite{wei2024q} by 1.13\% in accuracy with a minimal increase in parameters. 
These experiments demonstrate that our method significantly enhances the accuracy of BSNN while maintaining BSNN's efficiency.

\subsection{Ablation study}
In this section, we conduct ablation experiments to prove that the proposed AGMM mitigate the frequent weight sign flipping problem by regulating the gradient of BSNNs. And we evaluate AGMM-BSNN's energy-efficient advantages compared to vanilla BSNN and FP-SNN. All ablation experiments are performed on the CIFAR-100 dataset with the architecture of ResNet-19.

\subsubsection{Efficacy of AGMM}
\begin{figure}[tbp]
    \centering
    \subfigure[]{\includegraphics[width=0.49\linewidth]{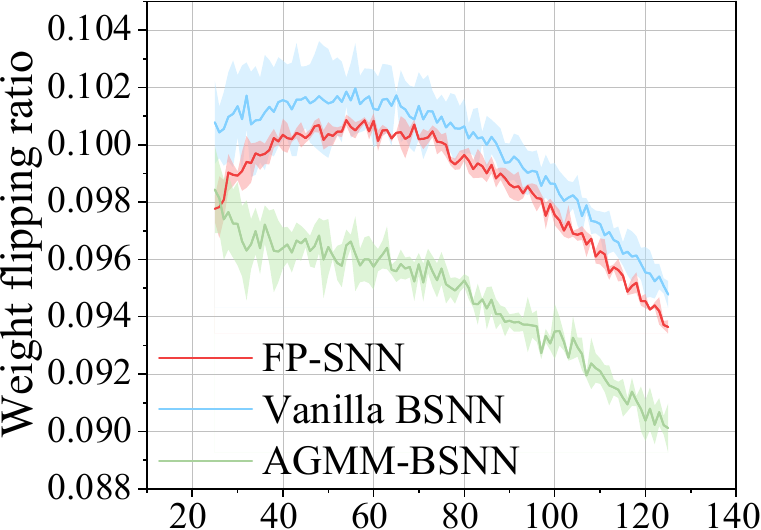}\label{fig:all_flip}}
    \subfigure[]{\includegraphics[width=0.50\linewidth]{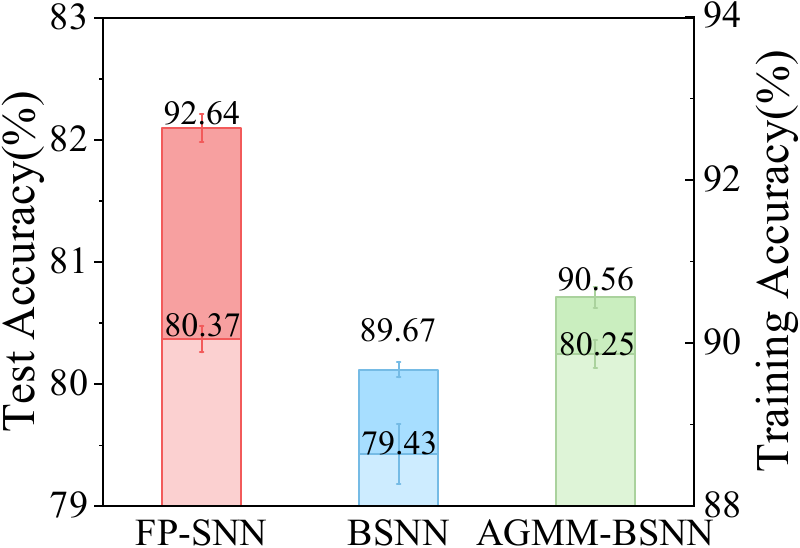}\label{fig:all_acc}}
    \caption{Weight sign flipping ratio and performance of FP-SNN, vanilla BSNN and AGMM-BSNN. The darker color represents the training accuracy, while the lighter color indicates the test accuracy.}
    \label{fig:all}
\end{figure}

\begin{figure}[tbp]
    \centering
    \subfigure[Vanilla BSNN]{\includegraphics[width=0.49\linewidth]{figs/baseline_loss.pdf}}
    \subfigure[AGMM-BSNN]{\includegraphics[width=0.49\linewidth]{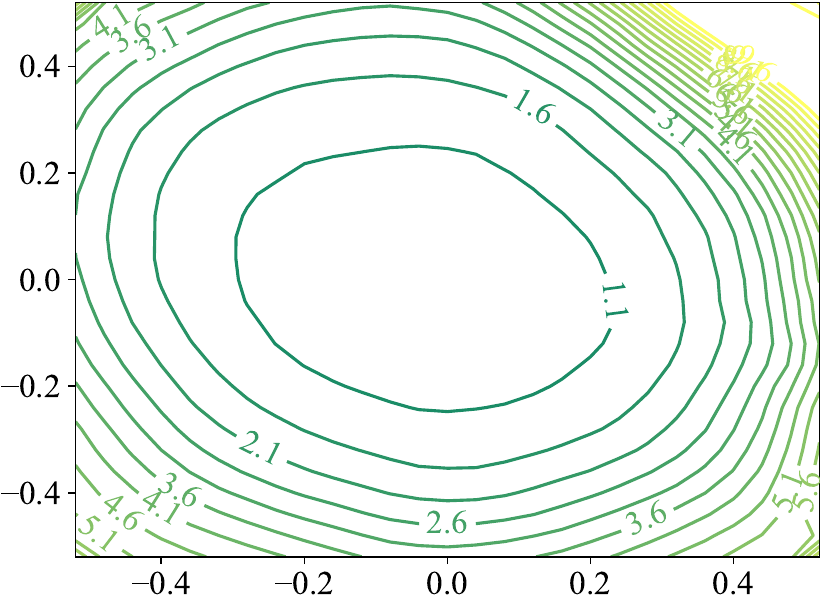}}
    \caption{The loss landscape of vanilla BSNN and AGMM-BSNN. AGMM-BSNN performs better in terms of generalization around the local minima.}
    \label{fig:agmm_loss}
\end{figure}

We first visualized the weight sign flipping ratio and performance between AGMM-BSNN, vanilla-BSNN, and FP-SNN to validate the efficacy of our proposed AGMM.
As shown in Fig.~\ref{fig:all_flip}, where the x-axis denotes the number of epochs, the flipping ratio of AGMM-BSNN is generally lower than that of the vanilla BSNN during the training process. Furthermore, Fig.~\ref{fig:all_acc} demonstrates that AGMM-BSNN significantly outperforms the vanilla BSNN in terms of accuracy. Although AGMM-BSNN's training accuracy is lower than that of FP-SNN, its test accuracy is comparable to FP-SNN, demonstrating the generalization capability of AGMM-BSNN.
We further inspect the 2D landscapes of AGMM-BSNN and vanilla BSNN around their local minima to demonstrate why AGMM-BSNN generalizes better than vanilla BSNN. As shown in Fig.~\ref{fig:agmm_loss}, AGMM-BSNN converges to a lower loss compared to vanilla BSNN, with a flatter convergence region. Moreover, AGMM-BSNN exhibits lower loss in areas distant from the convergence region. These observations indicate that AGMM enhances BSNN's convergence and improves its generalization performance. Based on the aforementioned results, it can be deduced that AGMM enhances BSNN's performance by decreasing weight sign flipping frequency, hence improving convergence.

\subsubsection{Efficiency of AGMM}
\begin{figure}
    \centering
    \includegraphics[width=0.8\linewidth]{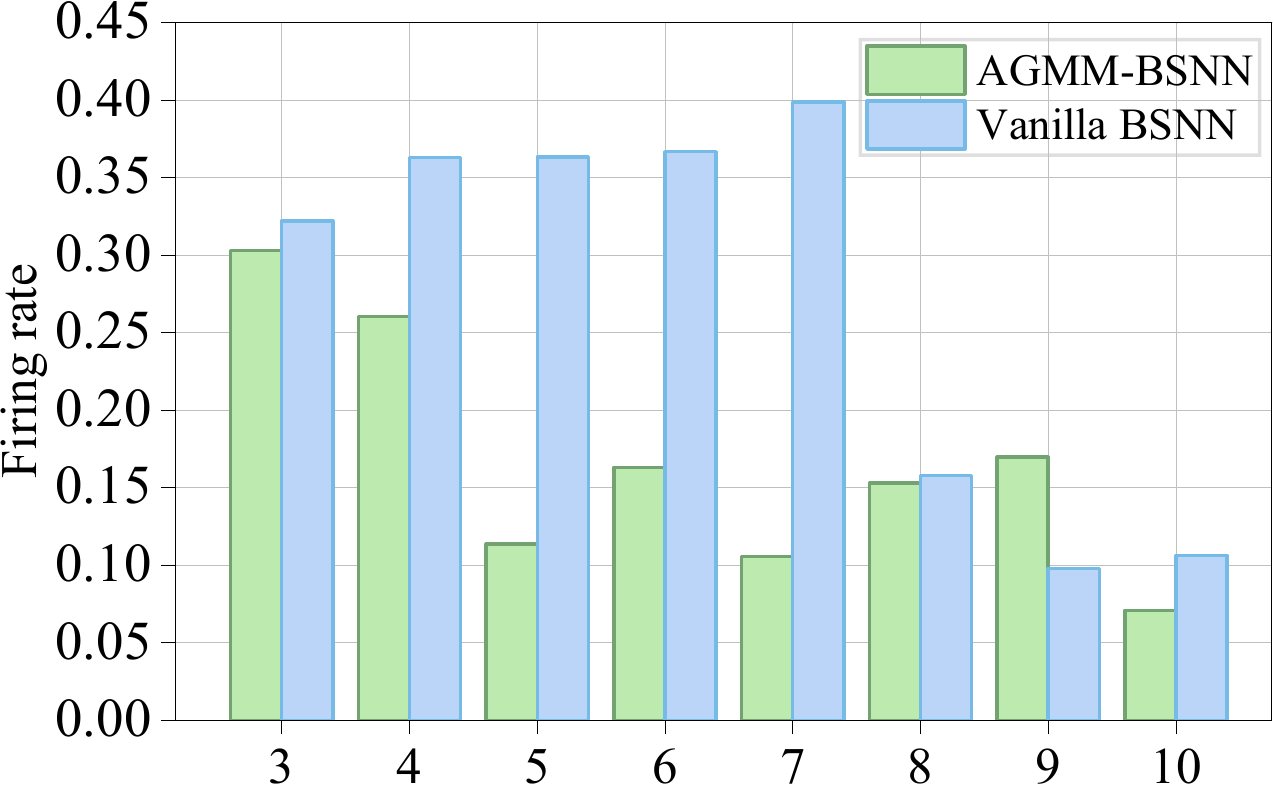}
    \caption{Firing rate in convolution layer of ResNet-19.}
    \label{fig:fire_rate}
\end{figure}

Although the proposed AGMM introduces a small number of learnable parameters, it does not compromise the efficiency of the BSNN. 
To demonstrate this, we first compare the firing rates of AGMM-BSNN and vanilla-BSNN. As shown in Fig.~\ref{fig:fire_rate}, since AGMM compresses the input to the spiking neurons during the forward pass via the sigmoid function, resulting in a lower firing rate for AGMM-BSNN. This phenomenon is evident in the middle layers of the network.
Additionally, we calculate the SOPs and power consumption of AGMM, following the methodology outlined in \cite{yao2024spike}, as shown in Tab.~\ref{energy}. While our method introduces additional parameters compared to the vanilla BSNN, the reduced firing rate results in lower SOP and energy consumption for AGMM, as compared to the vanilla BSNN.
\begin{table}[]
\centering
\def\arraystretch{1.2}
\begin{tabular}{c|cccc}
\hline
\multirow{2}{*}{Model} & Size          & SOPs           & MACs           & Energy         \\
                       & (MB)          & (G)            & (G)            & (mJ)           \\ \hline
AGMM-BSNN              & 2.46          & {0.045} & 0.028          & \textbf{0.064} \\
Vanilla BSNN           & {2.31} & 0.056          &{0.027} & 0.066          \\
FP-SNN                 & 50.46         & 0.750          &{0.027} & 0.330          \\ \hline
\end{tabular}
\caption{Energy estimation.}
\label{energy}
\end{table}

\section{Conclusion}
In this paper, we provide a comprehensive analysis of the weight sign flipping problem during the learning process of BSNNs and introduce an adaptive gradient modulation mechanism to address this challenge. Our theoretical analysis shows that the frequency of weight sign flipping is positively correlated with the mean and variance of gradients during training. To mitigate this issue, AGMM adaptively modulates the gradient magnitude, thereby reducing the mean and variance metrics in BSNNs. Extensive experiments on both static and neuromorphic datasets demonstrate that AGMM-BSNN consistently outperforms existing quantized SNN methods. These results underscore the effectiveness of the AGMM approach in enhancing the performance of BSNNs while maintaining their inherent energy efficiency advantages, making them more suitable for deployment in resource-constrained environments. In our future work, we will explore extending AGMM to complex large-scale SNN architectures, such as large spiking language models, and investigating its applicability to more complex tasks.

\section{Acknowledgments}
This work was supported in part by the National Natural Science Foundation of China under grant U20B2063, 62220106008, and 62106038, the Sichuan Science and Technology Program under Grant 2024NSFTD0034 and 2023YFG0259, the Open Research Fund of the State Key Laboratory of Brain-Machine Intelligence, Zhejiang University (Grant No.BMI2400020).

\bibliography{aaai25}

\end{document}